# Does deep learning always outperform simple linear regression in optical imaging?

**SHUMING JIAO,[1] YANG GAO,[1] JUN FENG,[1] TING LEI,[1,2] AND XIAOCONG YUAN[1,*]**

[1]*Nanophotonics Research Center, Shenzhen University, Shenzhen, Guangdong, China*
[2]*leiting@szu.edu.cn*
[*]*xcyuan@szu.edu.cn*

**Abstract:** Deep learning has been extensively applied in many optical imaging problems in recent years. Despite the success, the limitations and drawbacks of deep learning in optical imaging have been seldom investigated. In this work, we show that conventional linear-regression-based methods can outperform the previously proposed deep learning approaches for two black-box optical imaging problems in some extent. Deep learning demonstrates its weakness especially when the number of training samples is small. The advantages and disadvantages of linear-regression-based methods and deep learning are analyzed and compared. Since many optical systems are essentially linear, a deep learning network containing many nonlinearity functions sometimes may not be the most suitable option.

## 1. Introduction

In recent years, deep learning receives much attention in many research fields including optical design [1,2] and optical imaging [3]. In previous works, deep learning has been extensively applied for many optical imaging problems including phase retrieval [4-7], microscopic image enhancement [8-9], scattering imaging [10-11], holography [12-18], single-pixel imaging [19,20], super-resolution [21-24], Fourier ptychography [25-27], optical interferometry [28,29], wavefront sensing [30,31], and optical fiber communications [32].

Despite the success, deep learning has its own limitations and drawbacks, like any other approach [33]. For example, a huge number of training samples is usually required to train a deep neural network, which may not be always available in practical applications. The optimization of connection weights in the network with many training samples requires a considerable amount of computational cost. The design of network structure and the tuning of network parameters are often implemented empirically and intuitively with weak explainability. A deep neural network trained and tested for one category of samples may fail to work when it is generalized to other different testing samples. In fact, it is likely that deep learning may perform worse than other machine-learning (or non-machine-learning) methods in certain application scenarios. In previous works, the deficiencies of deep learning in solving optical imaging problems, compared with other methods, were seldom investigated.

In recent works [34,35], deep learning has been employed to address the problems of attacking a random-phase-encoded optical cryptosystem [34] and blind reconstruction for single-pixel imaging [35]. A random-phase-encoded optical cryptosystem is a coherent imaging system with multiple diffractive optical elements such as lens and random phase masks. The input plaintext light field is sequentially modulated by each phase mask in the forward propagation and it is finally transformed to a ciphertext light field as the system output. The objective of attacking a optical cryptosystem is to recover the input image from the given output light field if the encoding of all the phase masks is unknown. In single-pixel imaging [35, 36], the object image is sequentially illuminated by different structured light intensity patterns and the total light intensity of the entire object scene is recorded by a single-pixel detector for each pattern. Finally, the object image can be computationally reconstructed when both the illumination patterns and single-pixel intensity sequence are known. However, in a blind reconstruction [35], the objective is to recover the object image from the intensity sequence when the encoding of all the illumination patterns is unknown.

The two systems [34,35] stated above are both linear and can be regarded as a black box when the encoding of elements (phase masks or illumination patterns) is unknown. The random-phase-encoded optical cryptosystem is usually coherent while the single-pixel imaging is usually incoherent. In the previous work [37], it is shown that a multiple-phase-mask diffractive system and a single-pixel imaging system are similar from several aspects such as performing optical pattern recognition. In the previous works [34,35], each system is modeled by a different deep learning network optimized with many pairs of input and output training samples. Then the input image can be predicted by the network from an arbitrary given output. Since there is a linear relationship between the input and output of the two optical imaging systems mathematically, we point out that simple linear-regression-based methods can produce the same results as deep learning. A linear regression scheme can recover the object image more efficiently than deep learning for these two problems in some extent. The advantages and disadvantages of linear-regression-based methods and deep learning are analyzed and compared.

This paper is structured as follows. The linear regression model is described in Section 2. The two black-box optical imaging problems, i.e. attacking a random-phase-encoding-based optical cryptosystem and blind reconstruction in single-pixel imaging, are described in Section 3 and Section 4. The results and discussions about the comparison between linear-regression-based methods and deep learning are given in Section 5. A final conclusion is made in Section 6.

## 2. Linear regression model

For a linear optical imaging system, both the input X and output Y can be denoted by a column vector $X = [x_1 \; x_2 \; L \; x_M]$ and $Y = [y_1 \; y_2 \; L \; y_N]$. It is assumed that the input X has totally M pixels and the output Y has totally N pixels. The relationship between Y and X can be modeled as a matrix multiplication $Y=WX$, given by Eq. (1). The weighting matrix W consisting of $N \times M$ elements can be employed to model a black-box optical system.

$$\begin{bmatrix} y_1 \\ y_2 \\ M \\ y_N \end{bmatrix} = \begin{bmatrix} w_{11} & L & w_{1M} \\ M & O & M \\ w_{N1} & L & w_{NM} \end{bmatrix} \begin{bmatrix} x_1 \\ x_2 \\ M \\ x_M \end{bmatrix} \qquad (1)$$

When a large set of training samples (many pairs of X and Y) are available, the elements in the matrix W can be estimated by optimization if they are not given. The elements in W can be iteratively optimized with a gradient descent algorithm. Initially, all the elements in W are set to be random values. Then each element in W can be updated in the following way [38] based on the gradient descent for each training sample: $w'_{nm} = w_{nm} + r(y_n - y'_n)x_m \; (1 \leq m \leq M, 1 \leq n \leq N)$, where $y'_n$ denotes the actual output generated from the input of one training sample by multiplying the current W, $y_n$ denotes the target output of one training sample and r denotes the pre-defined learning rate. If the input values are complex-amplitude instead of real intensities, the algorithm needs to be slightly modified as: $w'_{nm} = w_{nm} + r(y_n - y'_n)conj(x_m) \; (1 \leq m \leq M, 1 \leq n \leq N)$, where $conj(\;)$ denotes the conjugate of a complex value. After many iterations, the adaptively optimized W matrix multiplied with a given input will yield an output close to the target one. A linear regression model can be considered as one-layer fully connected neural network without nonlinear activation functions, shown in Fig. 1. In a true fully connected neural network shown in Fig. 1(b), both linear connections and nonlinear activation functions are densely interconnected in the form of multiple cascaded layers between the network input and output. Modern deep learning networks such as the ones proposed in the previous works [34,35] shown in Fig. 3 and Fig. 5 usually have even more complicated structures than a fully connected neural network. Compared with a deep learning network, a linear regression model has very low complexity.

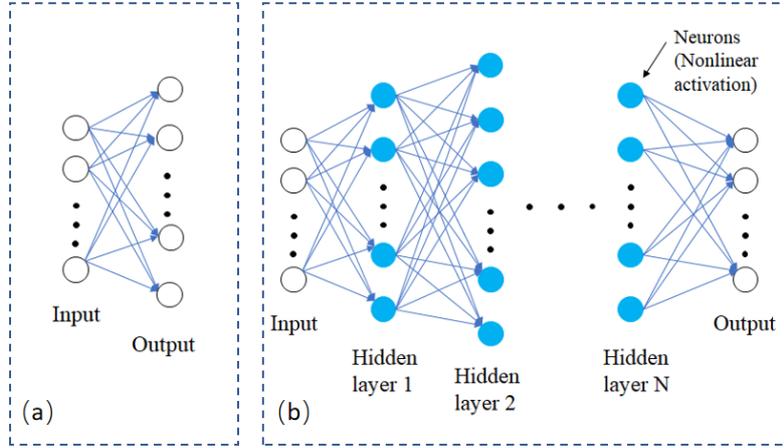

Fig. 1. (a) Linear regression model; (b) A fully connected neural network.

## 3. Problem 1: attacking a random-phase-encoding-based optical cryptosystem

As proposed in many previous works [34,39,40], an optical image encryption system can be constructed with an optical setup consisting of cascaded lens and random phase masks. Typical examples include Double Random Phase Encryption (DRPE) and Triple Random Phase Encryption (TRPE) [34]. In this work, the one with a more complicated structure, i.e. a TRPE system, is considered and its optical setup is shown in Fig. 2. In a TRPE system, the pixel intensities of the input light field represent the plaintext image O. Then the input light field is optically Fourier transformed and inverse Fourier transformed with a double-lens 4f setup. The light field in the output plane becomes the ciphertext C. The plaintext image can be decrypted from the ciphertext with the same setup by backward light field propagation. Three random phase masks $R_1$, $R_2$ and $R_3$ are placed in the input plane, the Fourier plane and the output plane respectively. The pixel values of all the phase masks are encoded as random phases between $[0\ 2\pi]$. The three phase masks serve as the encryption and decryption key. The mathematical model of TRPE encryption and decryption is given by Eqs. (2) and (3).

$$C = IFT\left[FT\left(O \cdot \exp(i \cdot R_1)\right) \cdot \exp(i \cdot R_2)\right] \cdot \exp(i \cdot R_3) \quad (2)$$

$$O = IFT\left\{FT\left[C \cdot \exp(-i \cdot R_3)\right] \cdot \exp(-i \cdot R_2)\right\} \cdot \exp(-i \cdot R_1) \quad (3)$$

where FT and IFT denotes Fourier transform and inverse Fourier transform, O denotes the input plaintext image and C denotes the encrypted light field (ciphertext).

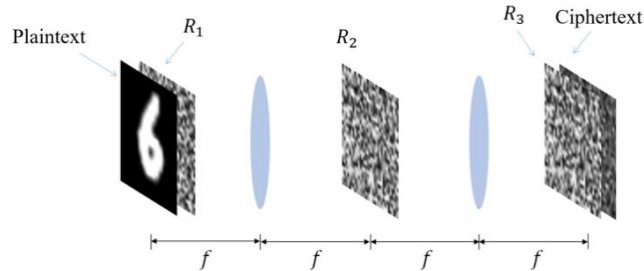

Fig. 2. Optical setup of a triple random phase encryption (TRPE) system.

Ideally, the plaintext image O cannot be recovered from the ciphertext C if the key is not known and the information security is protected in this way. However, the encryption system can be cracked by a known-plaintext attack (KPA) if the attacker collects an adequate number of plaintext-ciphertext pairs. In KPA, the objective is to recover the plaintext O from the corresponding ciphertext C without knowing $R_1$, $R_2$ and $R_3$. The entire system is linear and the ciphertext can be regarded as the input vector X and the plaintext can be regarded as the output Y in the linear regression model described in Section 2. The two-dimensional matrices C and O can be rearranged as one-dimensional vectors X and Y. Consequently, a KPA to a TRPE system can be implemented with complex-amplitude linear regression (CLR), in addition to deep learning. In the previous work [34], the deep learning network structure shown in Fig. 3, referred to as DecNet, was employed for the KPA. In this work, CLR is compared with DecNet for the same KPA attack to a TRPE system.

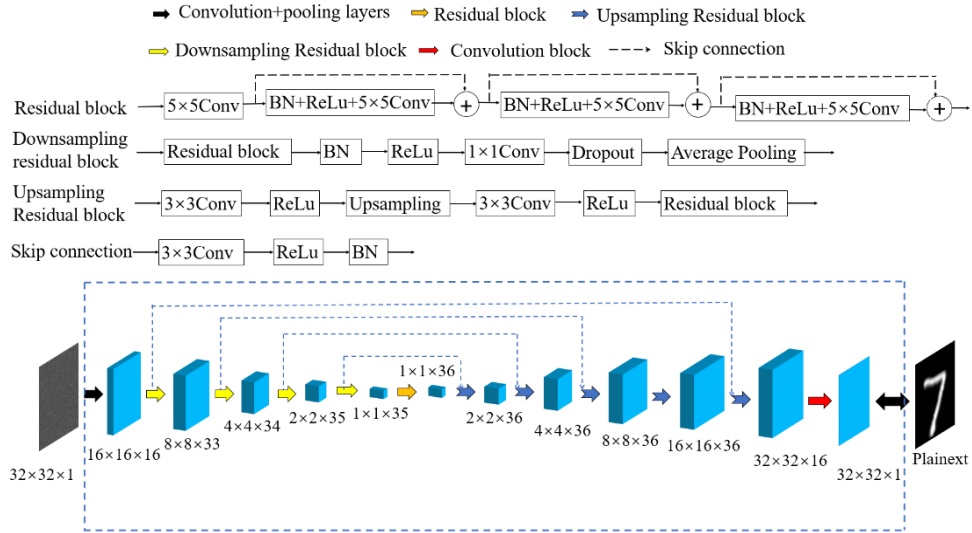

Fig. 3. Deep learning network for attacking a TRPE system proposed in the previous work [34] (DecNet).

## 4. Problem 2: blind reconstruction in single-pixel imaging

In single-pixel imaging (SPI), the light intensity is recorded by a sensor containing only one single pixel, instead of a pixelated sensor array. A typical optical setup for a SPI system is shown in Fig. 4.

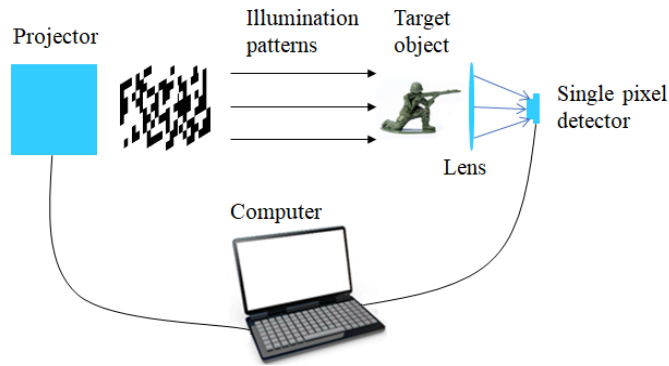

Fig. 4. Optical setup of a single-pixel imaging system.

The two-dimensional object image $O(x, y)$ is sequentially illuminated by N varying two-dimensional structured light patterns $P_n(x, y)(1 \leq n \leq N)$ and a single-pixel intensity sequence $I_n (1 \leq n \leq N)$ will be recorded. Mathematically, each element in $I_n$ is the inner product between $O(x, y)$ and each pattern in $P_n(x, y)$. The object image $O(x, y)$ can be computationally reconstructed when both the illumination pattern sequence $P_n(x, y)$ and the recorded intensity sequence $I_n$ are known. It is assumed that the total number of pixels in $O(x, y)$ and $P_n(x, y)$ is M. The sampling ratio S can be defined as N/M. In single-pixel imaging, various kinds of algorithms can be employed to reconstruct $O(x, y)$ from $P_n(x, y)$ and $I_n$ [41]. However, all the illumination patterns $P_n(x, y)$ are required to be known in these reconstruction algorithms. It is usually easier to reconstruct a high-quality object image when the sampling ratio S is higher. A blind reconstruction in SPI by deep learning was attempted in the previous work [35], where the object image $O(x, y)$ is recovered from only $I_n$ when the patterns $P_n(x, y)$ are not given. The blind reconstruction in SPI is favorable for some applications such as scattering imaging [35, 42]. It is assumed that multiple pairs of different object images and single-pixel intensities are given for the fixed illumination patterns, which can be used as training samples in deep learning.

The blind reconstruction in SPI essentially contains two steps: (a) Recovery of the unknown illumination patterns $P_n(x, y)$ from the training samples. This is similar to the KPA in random-phase-encoding-based optical encryption described in Section 3; (b) Object image reconstruction in SPI from a given $I_n$ and the estimated $P_n(x, y)$ obtained in Step (a). Step (a) is most critical and it is the key in the blind reconstruction. Once the patterns can be approximately estimated and recovered in Step (a), Step (b) is simply conventional image reconstruction in SPI and there are many different methods proposed in the past [41]. The image reconstruction quality in Step (b) mainly depends on the accuracy of the estimated illumination patterns by linear regression in Step (a). The deep learning approach in the previous work [35] is end-to-end and both two steps are realized within the network shown in Fig. 5.

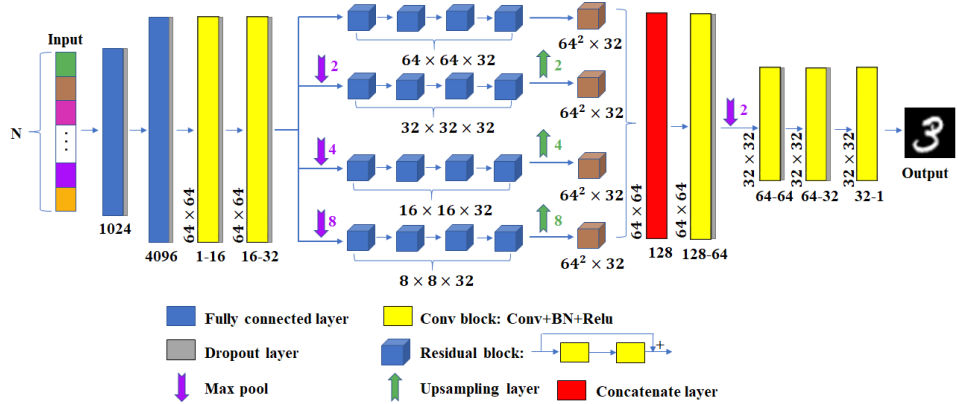

Fig. 5. Deep learning network for blind image reconstruction in SPI proposed in the previous work [35] (Wang's Net).

In SPI, $O(x, y)$ and $I_n$ have a linear mathematical relationship. The M pixels in $O(x, y)$ can be rearranged as the one-dimensional input vector X in Equation (1) and $I_n$ is equivalent to the output vector Y in Eq. (1). All the N illumination patterns $P_n(x, y)$ will jointly constitute the weighting matrix W in Eq. (1) and each pattern corresponds to one row in W. Consequently, the unknown illumination patterns can be recovered from the training samples by linear regression for Step (a). Then a compressive sensing scheme with total variation minimization [41,43,44] can be employed for image reconstruction in Step (b). No training samples are required for compressive sensing reconstruction since it is not a machine learning process. Our proposed scheme is referred to as "Linear Regression + Compressive Sensing (LRCS)". The LRCS scheme is compared with the deep learning network proposed in the previous work [35], referee to as Wang's Net. It shall be noted that no linear regression is performed to recover the illumination patterns in the previous work [35], even though compressive sensing is adopted for image reconstruction by assuming the illumination patterns are already known.

## 5. Results and discussion

*5.1 Attacking a random-phase-encoding-based optical cryptosystem*

A complex-amplitude linear regression (CLR) is performed to crack a TRPE optical cryptosystem. For comparison, a DecNet [34] is constructed and the corresponding cracking results are obtained as well. The size of plaintext image, ciphertext and random phase masks is $32 \times 32$ pixels. Plaintext images are randomly selected from the number-digit images in the MNIST dataset [45], the fashion product images in the Fashion-MNIST dataset [46] and natural object images in the CIFAR-100 dataset [47]. The color images in the CIFAR-100 dataset are converted to grayscale images. The output ciphertext light fields corresponding to the plaintext images are generated from a simulated TRPE system. In the training, plaintext images are used as the target output and complex-amplitude ciphertexts are used as the input for both CLR and DecNet. Various number of training samples are attempted: 50, 100, 200, 500, 2000 and 5000. In addition, 200 samples randomly selected from each dataset different from the training samples are employed to test the attacking capability of the CLR and DecNet after training. The peak-signal-to-noise-ratio (PSNR) between the original plaintext image and the recovered result from the ciphertext by these two methods is employed to evaluate their performance.

In CLR, the learning rate is 0.01 for the MNIST dataset and 0.001 for the Fashion-MNIST dataset and the CIFAR-100 dataset. The number of iterations is set to be 300. In DecNet, the learning rate is 0.0001 and the number of epochs is 20 for all the datasets. The results of our complex-amplitude linear regression are compared with the ones using deep learning [34] in Table 1, Table 2, Table 3, Fig. 6 and Fig. 7. The training time of CLR is 825 seconds for 100 training samples and 4155 seconds for 500 training samples in a Matlab R2018a environment with Intel(R) Core(TM) i5-8400U CPU (2.80 GHz) and 8GB RAM. The training time of DecNet is 54 seconds for 100 training samples, 203 seconds for 500 training samples, 804 seconds for 2000 training samples and 2016 seconds for 5000 training samples under the Keras framework in a Python 3.5 environment. The training time of DecNet is generally shorter than CLR. The inference time of a trained model for predicting a plaintext from a ciphertext is both within 0.1 second in CLR and DecNet.

For the MNIST dataset and Fashion-MNIST dataset, it can be observed that the performance of both methods will be improved as the number of training samples increases. However, CLR performs much better than DecNet when the number of training samples is small (e.g. from 50 to 500). The DecNet can only yield satisfactory output results when the number of training samples is at least 2000 or 5000. The results from CLR with 200 training samples is close to the results from DecNet with 5000 samples. Evidently, CLR has significant advantages compared with DecNet in attacking a TRPE system when the number of training samples is inadequate. It can be observed from Fig. 6 that the recovered MNIST images by DecNet are contaminated with stripe noise and the recovered Fashion-MNIST images by DecNet are heavily blurred when the number of training samples is small. Theoretically, it is possible for a deep learning network like DecNet to accurately model a linear system. However, the network may be overfitted at a local optimal point in the training when the number of training samples is small. Since the global optimal solution is not reached, the network may yield unfavorable prediction results for the testing images.

For the CIFAR-100 dataset, the performance of CLR is close to that for the Fashion-MNIST dataset and the original plaintext images can be recovered with acceptable visual quality. But DecNet completely fails to recover the plaintext images even when the number of training samples is adequate (up to 5000). The CIFAR-100 dataset contains complicated natural images from 100 different categories of objects. The MNIST dataset (or Fashion-MNIST dataset) only contains ten different kinds of simple number digit images (or fashion product images) and all the images have black background. It is easier for DecNet to extract common features from the MNIST or Fashion-MNIST images and perform plaintext recovery from a ciphertext based on these features. But it is difficult to extract common features from the diversified CIFAR-100 images for DecNet. CLR does not extract high-level features from the images so its performance will not vary a lot for different datasets.

In the previous work [34], the DecNet can still work when only ciphertext intensities are available as the network input, instead of both ciphertext intensities and phases. However, CLR will not work in this situation since the input-output relationship is no longer linear. This is one major limitation of CLR compared with DecNet.

**Table 1. Recovered plaintext image quality with CLR and DecNet (MNIST dataset)**

| Number of training samples K | CLR | DecNet |
| --- | --- | --- |
| 50 | 17.4439 dB | 11.5822 dB |
| 100 | 19.4437 dB | 11.6669 dB |
| 200 | 22.4502 dB | 12.0497 dB |
| 500 | 27.2989 dB | 13.0242 dB |
| 2000 | ---- | 13.6974 dB |
| 5000 | ---- | 21.3453 dB |

**Table 2. Recovered plaintext image quality with CLR and DecNet (Fashion-MNIST dataset)**

| Number of training samples K | CLR | DecNet |
| --- | --- | --- |
| 50 | 15.8603 dB | 11.1401 dB |
| 100 | 17.7190 dB | 12.1886 dB |
| 200 | 19.8130 dB | 15.3303 dB |
| 500 | 22.9465 dB | 17.4687 dB |
| 2000 | ---- | 19.3498 dB |
| 5000 | ---- | 20.6077 dB |

**Table 3. Recovered plaintext image quality with CLR and DecNet (CIFAR-100 dataset)**

| Number of training samples K | CLR | DecNet |
| --- | --- | --- |
| 50 | 15.8243 dB | 7.4102 dB |
| 100 | 17.3781 dB | 8.3004 dB |
| 200 | 18.9631 dB | 10.1093 dB |
| 500 | 21.5758 dB | 11.8484 dB |
| 2000 | ---- | 12.7018 dB |
| 5000 | ---- | 10.7213 dB |

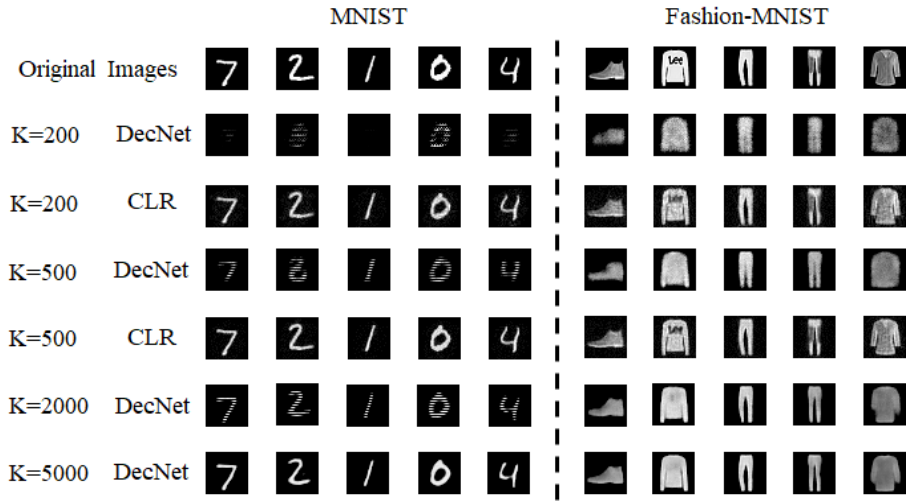

Fig. 6. Comparison of recovered plaintext image results for a TRPE system with CLR and DecNet (MNIST dataset and Fashion-MNIST dataset).

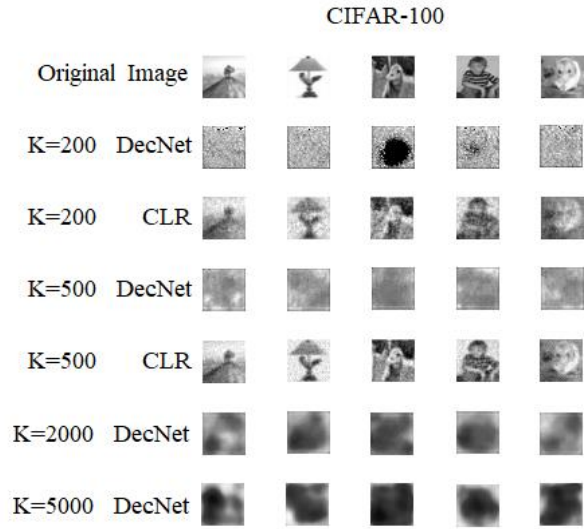

Fig. 7. Comparison of recovered plaintext image results for a TRPE system with CLR and DecNet (CIFAR-100 dataset).

The similarities and differences between the proposed CLR scheme and the previously proposed DecNet for attacking a TRPE system are summarized in Table 4.

Table 4. Similarities and differences between CLR and DecNet for attacking a TRPE system

|  | Complex-amplitude Linear Regression (CLR) | DecNet |
|---|---|---|
| Similarity | Predict the plaintext image from the ciphertext light field for a TRPE system (or other similar optical cryptosystems) after the model is trained with a certain number of training samples | |
| Difference | (1) Work with a small number of training samples<br>(2) Simple explicit model with few parameters<br>(3) Not work for intensity-only ciphertext<br>(4) Work for both simple images with black background and complicated natural images | (1) Only work with a large number of training samples<br>(2) Complicated black-box model with many parameters for tuning<br>(3) Work for intensity-only ciphertext<br>(4) Work for simple images with black background, not work for complicated images |

*5.2 Blind reconstruction in single-pixel imaging*

In the simulation, the size of object image and each illumination pattern is $32 \times 32$ pixels. The pixel intensity values in each illumination pattern are randomly distributed between 0 and 1. Four different numbers of illuminations, N=51, N=205, N=410 and N=1024 corresponding to four different sampling ratios S=0.05, S=0.2, S=0.4 and S=1, are attempted. Various number of training images and 200 testing images are randomly selected from the MNIST dataset and the CIFAR-100 dataset. The single-pixel intensity values can be obtained based on the SPI model described in Section 4. Both our proposed "linear regression + compressive sensing" (LRCS) scheme and Wang's Net proposed in the previous work [35] are implemented to recover the original object image. The optimization solver for the compressive-sensed image reconstruction in LRCS is based on the work [43] for the MNIST dataset and based on the work [41] for the CIFAR-100 dataset. In the linear regression step of LRCS, the learning rate is set to be 0.01 and the number of iterations is 300 for all the cases. In the training of Wang's Net, the learning rate is set to be 0.00001 and the number of epochs is 20 for all the cases. The average PSNR of the blindly reconstructed images from the simulated single-pixel intensity values for the 200 testing samples is presented in Table 5, Table 6, Table 7 and Table 8. Some examples of the reconstructed images are shown in Fig. 8 and Fig. 9.

For the results of the MNIST dataset in Table 5, Table 6 and Fig. 8, it can be observed that the performance of LRCS will be enhanced as the sampling ratio increases and the number of training samples increases. The performance of deep learning will be significantly enhanced as the number of training samples increases but it will not be necessarily improved as the sampling ratio increases. At a very low sampling ratio S=0.05, the reconstructed images by LRCS are very heavily degraded but most reconstructed images by deep learning still have acceptable visual quality when the number of training samples are adequate. Since deep learning can extract some high-level common features from the training images, the test object image can still be well recovered from these features when the sampling ratio is very low. On the other hand, the feature extraction and reconstruction may cause more unpredicted errors in the recovered images when the dimension of input data is higher. So the quality of recovered images will not always be worse at a lower sampling ratio and better at a higher sampling ratio for the deep learning approach.

**Table 5. Blindly reconstructed image quality for SPI with LRCS (MNIST dataset)**

| Number of training samples K | S=0.05 | S=0.2 | S=0.4 | S=1 |
| --- | --- | --- | --- | --- |
| 50 | 14.2419 dB | 16.8031 dB | 17.5811 dB | 18.3476 dB |
| 100 | 15.1527 dB | 18.7134 dB | 20.0936 dB | 21.3368 dB |
| 200 | 15.5376 dB | 20.1990 dB | 22.1549 dB | 24.2062 dB |
| 500 | 15.3003 dB | 21.0111 dB | 23.6042 dB | 26.4277 dB |

**Table 6. Blindly reconstructed image quality for SPI with Wang's Net (MNIST dataset)**

| Number of training samples K | S=0.05 | S=0.2 | S=0.4 | S=1 |
| --- | --- | --- | --- | --- |
| 50 | 11.9738 dB | 11.6227 dB | 12.2526 dB | 11.5919 dB |
| 100 | 12.8692 dB | 12.6286 dB | 12.8774 dB | 12.6637 dB |
| 200 | 13.0954 dB | 13.2059 dB | 13.2644 dB | 13.1486 dB |
| 500 | 13.2826 dB | 13.3295 dB | 13.3089 dB | 13.2944 dB |
| 2000 | 15.5567 dB | 14.7390 dB | 17.1544 dB | 15.2462 dB |
| 5000 | 17.3145 dB | 19.3743 dB | 20.1935 dB | 19.7935 dB |

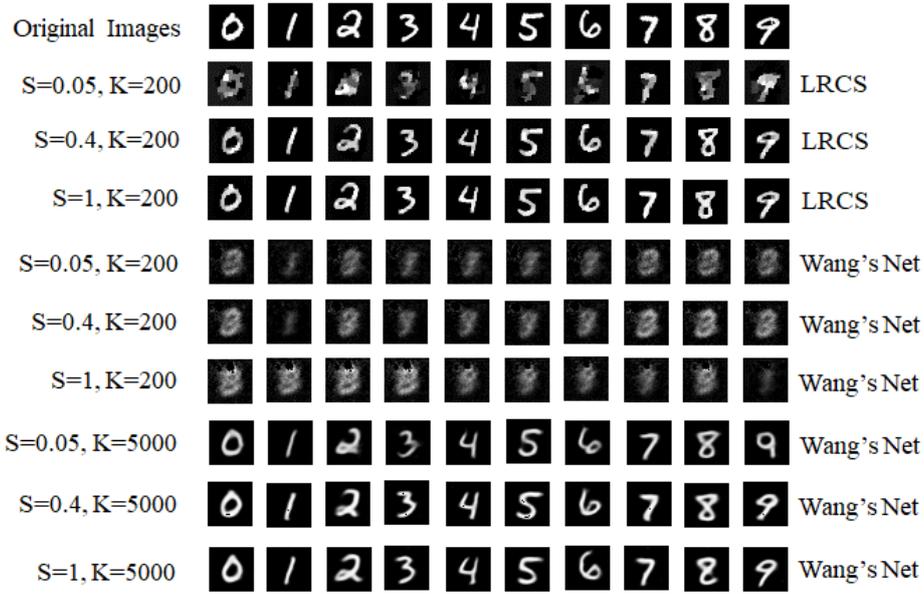

Fig. 8. Comparison of reconstructed image results for a SPI system with LRCS and deep learning in the simulation (MNIST dataset).

Fig. 8 shows that some reconstructed images by LRCS have quality degradation but the shapes of the digits match with the original groundtruth MNIST images. On the other hand, some reconstructed images by Wang's Net can be noise-free but the digits have distorted shapes, which will cause a lower PSNR. From the results, it can be observed that the recovered image quality of LRCS with 200 training samples is comparable with the ones using deep learning with 5000 samples, except when the sampling ratio is very low.

From the results of the CIFAR-100 dataset in Table 7, Table 8 and Fig. 9, it can be observed that the performance of LRCS is not as good as the one for the MNIST dataset but the object image can still be reconstructed with high fidelity under proper conditions. On the other hand, Wang's Net fails to work for the complicated natural object images in the CIFAR-100 dataset, regardless of sampling ratios and number of training samples. Similar to the DecNet, it is also hard for Wang's Net to extract high-level common features from the CIFAR-100 images to reconstruct high-quality results. LRCS performs significantly better than Wang's Net for the CIFAR-100 dataset. Even though DecNet and Wang's Net both demonstrate poor performances for complicated natural object images, it is possible that other different deep learning models may work for these images in the two black-box imaging tasks.

Table 7. Blindly reconstructed image quality for SPI with LRCS (CIFAR-100 dataset)

| Number of training samples K | S=0.05 | S=0.2 | S=0.4 | S=1 |
| --- | --- | --- | --- | --- |
| 200 | 14.7771 dB | 16.4979 dB | 17.1860 dB | 17.3206 dB |
| 500 | 15.0334 dB | 17.6346 dB | 18.8980 dB | 19.2882 dB |
| 1000 | 14.9699 dB | 18.0890 dB | 20.1176 dB | 21.4797 dB |

Table 8. Blindly reconstructed image quality for SPI with Wang's Net (CIFAR-100 dataset)

| Number of training samples K | S=0.05 | S=0.2 | S=0.4 | S=1 |
| --- | --- | --- | --- | --- |
| 200 | 10.2119 dB | 10.8575 dB | 10.6441 dB | 10.6401 dB |
| 500 | 10.9371 dB | 11.1875 dB | 11.0173 dB | 11.0095 dB |
| 1000 | 10.8714 dB | 11.1994 dB | 10.9367 dB | 10.9383 dB |
| 5000 | 12.6881 dB | 13.3186 dB | 12.2194 dB | 11.0790 dB |

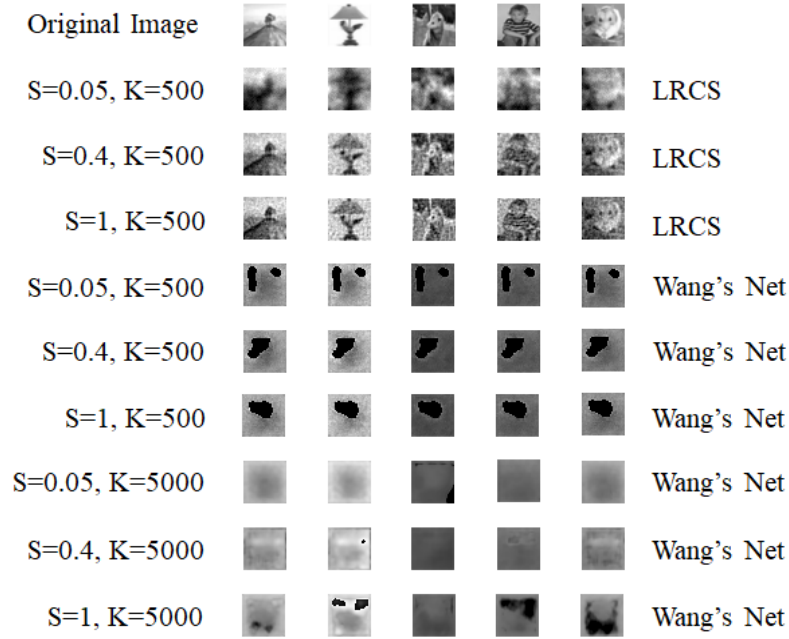

Fig. 9. Comparison of reconstructed image results for a SPI system with LRCS and deep learning in the simulation (CIFAR-100 dataset).

Table 9. Similarities and differences between LRCS and Wang's Net for blind reconstruction in SPI

|  | Linear Regression + Compressive Sensing (LRCS) | Deep learning (Wang's Net) |
| --- | --- | --- |
| Similarity | Blindly reconstruct the object image from the single-pixel intensity sequence at different sampling ratios for a SPI system after the model is trained with a certain number of training samples, when the illumination patterns are not known | |
| Difference | (1) Work with a small number of training samples<br>(2) Simple explicit model with few parameters<br>(3) Two-step, not end-to-end<br>(4) Poor performance at a very low sampling ratio<br>(5) Work for both simple images with black background and complicated natural images | (1) Only work with a large number of training samples<br>(2) Complicated model with many parameters for tuning<br>(3) End-to-end reconstruction<br>(4) Possibly reconstruct high-quality results at a very low sampling ratio<br>(5) Work for simple images with black background, not work for complicated images |

It takes about 725 seconds for 200 training samples and 1882 seconds for 500 training samples to train the linear regression part in the LRCS scheme (S=1), with the same hardware and software configuration in Section 5.1. When the sampling ratio S is lower, the training time will be reduced by being multiplied with S. In contrast, it takes about 790 seconds for 200 training samples, 1940 seconds for 500 training samples, 8060 seconds for 2000 training samples and 20591 seconds for 5000 training samples to train Wang's Net. The training time of Wang's Net will not be evidently reduced if the sampling ratio becomes lower. LRCS is generally more computationally efficient than Wang's Net in the training step. Wang's Net is less efficient than DecNet in the training because the latter one mainly consists of convolutional layers while the former one consists of several fully connected layers. The inference time of a trained Wang's Net to reconstruct a testing image is within 0.1 second. The reconstruction time of LRCS for a testing image will be around 0.3 to 0.5 second, which is relatively longer since the optimization step in compressive sensing requires certain computational cost. The similarities and differences between LRCS and deep learning from different perspectives are summarized in Table 9.

In this work, LRCS and Wang's Net are evaluated based on the experimentally recorded data as well. The SPI experiments are conducted using the optical setup shown in Fig. 10. Each object image is printed on a paper card and illuminated by the patterns projected by a JmGO G3 projector. The single-pixel intensity values are recorded by a Thorlabs FDS1010 photodiode detector and a NI USB-6216 data acquisition card. Totally ten different object images are tested in the experiment.

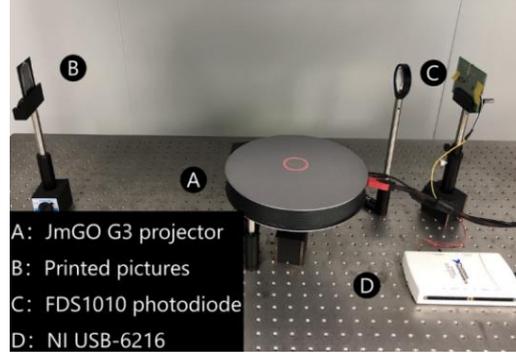

Fig. 10. Optical setup of our SPI experiment.

The reconstruction results with LRCS and Wang's Net from the experimentally recorded data are shown in Fig. 11. It is reported in the previous work [35] that deep learning is significantly more robust to the noise in the experimental data. Since the optical setup in this work is different from the one in the previous work [35], the type of noise and its strength can be different in the experiment. For example, no laser illumination is employed in this work and the speckle noise contamination will not occur. In our observation, the performances of both LRCS and Wang's Net are slightly degraded due to the extra experimental noise that do not appear in the simulated training data. But it is still evident that Wang's Net performs better than LRCS at a low sampling ratio and LRCS perform better than Wang's Net when the number of training samples is small.

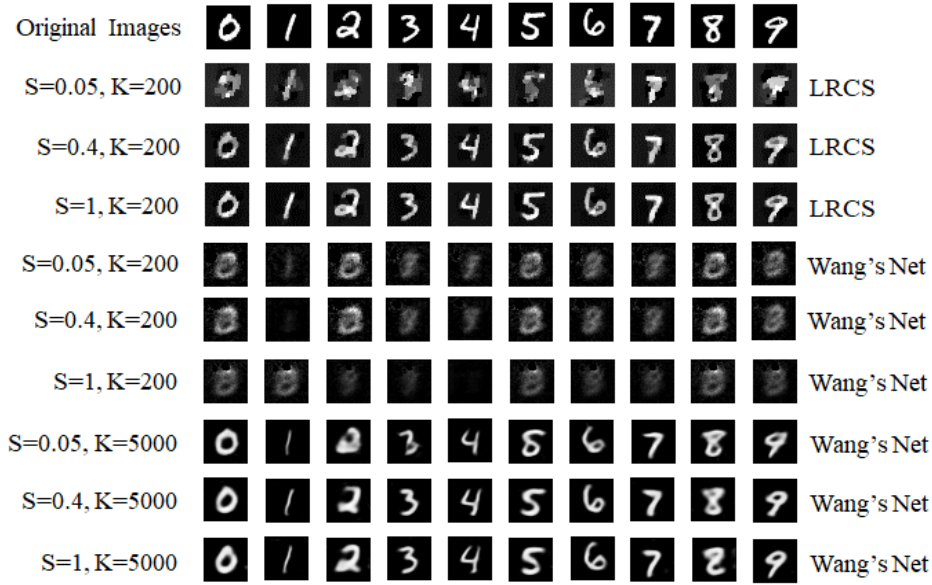

Fig. 11. Comparison of reconstructed image results with LRCS and Wang's Net for a SPI system based on the recorded data in real optical experiments.

## 6. Conclusion

In this work, we point out that linear-regression-based methods can be used to solve two black-box optical imaging problems that were previously addressed by deep learning approaches. For attacking a TRPE optical cryptosystem, a complex-amplitude linear regression (CLR) scheme is proposed. For the blind image reconstruction in a SPI system, a "linear regression + compressive sensing (LRCS)" scheme is proposed. In these two problems, linear-regression-based methods show some advantages than deep learning such as being applicable to a small number of training samples and complicated natural object images. Simulation and experimental results indicate that deep learning does not always outperform linear regression in this type of black-box optical imaging problems and each approach has its own advantages and disadvantages. Compared with linear regression, nonlinear deep learning models have advantages of recovering the original images when the given information is incomplete (e.g. only ciphertext intensity is known for TRPE or the sampling ratio is very low in SPI). The similarities and differences between linear-regression-based methods and deep learning are analyzed and summarized.

## Funding
National Natural Science Foundation of China (61805145, 11774240); Leading Talents Program of Guangdong Province (00201505); Natural Science Foundation of Guangdong Province (2016A030312010).

## Acknowledgement
We would like to thank Dr. Fei Wang, Mr. Han Hai and Prof. Wenqi He's assistance in this work.

## Disclosures
The authors declare no conflicts of interest.